\documentclass[letterpaper, 10 pt, conference]{ieeeconf}  % Comment this line out if you need a4paper

\IEEEoverridecommandlockouts                              % This command is only needed if 
\usepackage{graphicx}
\usepackage{multirow}
\usepackage{algorithm}
\usepackage{subcaption}
\usepackage{soul}
\usepackage{comment}
\usepackage[noend]{algpseudocode}
\usepackage{afterpage}

\title{\LARGE \bf
FlyNeRF: NeRF-Based Aerial Mapping for High-Quality 3D Scene
Reconstruction
}

\author{Maria Dronova, Vladislav Cheremnykh, Alexey Kotcov, Aleksey Fedoseev, and Dzmitry Tsetserukou%  
\thanks{The authors are with the Intelligent Space Robotics Laboratory, Skolkovo Institute of Science and Technology, Bolshoy Boulevard 30, bld. 1, 121205, Moscow, Russia}
\thanks{email: {(maria.dronova, vladislav.cheremnykh, alexey.kotcov, aleksey.fedoseev, d.tsetserukou})@skoltech.ru}}%
\begin{document}

\maketitle
\thispagestyle{empty}
\pagestyle{empty}

%%%%%%%%%%%%%%%%%%%%%%%%%%%%%%%%%%%%%%%%%%%%%%%%%%%%%%%%%%%%%%%%%%%%%%%%%%%%%%%%
\begin{abstract}

Current methods for 3D reconstruction and environmental mapping frequently face challenges in achieving high precision, highlighting the need for practical and effective solutions. In response to this issue, our study introduces FlyNeRF, a system integrating Neural Radiance Fields (NeRF) with drone-based data acquisition for high-quality 3D reconstruction. Utilizing unmanned aerial vehicle (UAV) for capturing images and corresponding spatial coordinates, the obtained data is subsequently used for the initial NeRF-based 3D reconstruction of the environment. Further evaluation of the reconstruction render quality is accomplished by the image evaluation neural network developed within the scope of our system. According to the results of the image evaluation module, an autonomous algorithm determines the position for additional image capture, thereby improving the reconstruction quality. 

The neural network introduced for render quality assessment demonstrates an accuracy of 97\%. Furthermore, our adaptive methodology enhances the overall reconstruction quality, resulting in an average improvement of 2.5 dB in Peak Signal-to-Noise Ratio (PSNR) for the 10{\%} quantile. The FlyNeRF demonstrates promising results, offering advancements in such fields as environmental monitoring, surveillance, and digital twins, where high-fidelity 3D reconstructions are crucial.

\end{abstract}

%%%%%%%%%%%%%%%%%%%%%%%%%%%%%%%%%%%%%%%%%%%%%%%%%%%%%%%%%%%%%%%%%%%%%%%%%%%%%%%%
\section{Introduction}

Autonomous navigation remains a significant challenge in the development of agents designed to operate in unknown environments such as cluttered rooms and facilities. Recently, Neural Radiance Field (NeRF) technology has found many applications in robotics, one of the most promising being the reconstruction of 3D scenes for further use in navigation, mapping, and path planning. However, despite its capabilities, the quality of NeRF renderings is determined by the quality of the input images, especially when the robot captures them at high speed, under challenging conditions, or from an unfavorable angle. Low-quality renders affect path planning and localization of the autonomous agent, leading to poor construction of the 3D map.

In the topic of exploration of unseen environments previous works have mainly focused on optimization of reward policies \cite{mutti2022unsupervised, savinov2019episodic} to maximize the survey area, finding a specific group of targets \cite{ramakrishnan2022poni, liang2021sscnav}, recognizing by images or language \cite{krantz2020navgraph, min2021film, nguyen2021look}. In our work, the main focus is assessing render quality and improving it with an autonomous agent using data from our neural network. Through this process, we obtain information about camera poses where image quality is below a certain threshold. This information is later used by the agent to provide new images. By iteratively refining the dataset with novel images, our proposed framework aims to improve the overall quality of the NeRF renderings and the accuracy of the associated 3D reconstructions.

\begin{figure}
      \centering
      \includegraphics[width=1.0\linewidth]{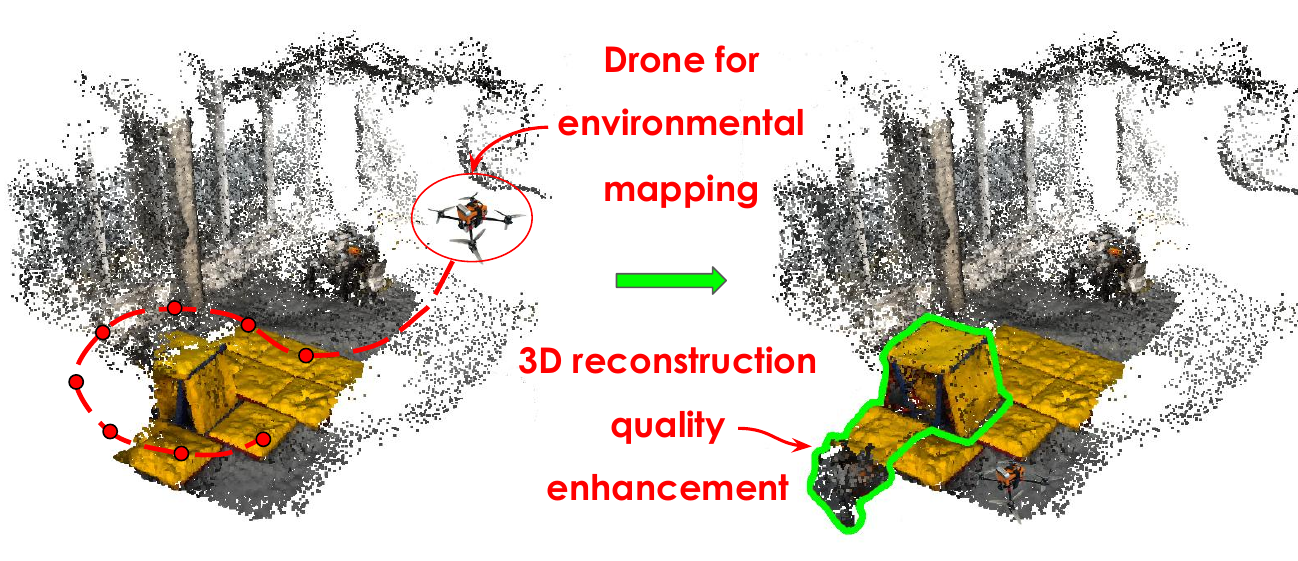}
      \caption{FlyNeRF system during the mission. The dashed red line with dots represents the trajectory executed by the drone and positions for additional image capture. The green area signifies the improvement in the quality of the reconstruction.}
      \label{teaser}
   \end{figure}

In summary, we present FlyNeRF (Fig.~\ref{teaser}), a novel system utilizing UAV to explore an unseen scene to collect and refine data for training a NeRF model. Our method uses a neural network-based quality assessment approach to identify images of poor quality based on Structural Similarity Index (SSIM) and Peak Signal-to-Noise Ratio (PSNR) metrics. These identified images, which are indicative of poor camera poses or environmental conditions, are then used to generate the list of positions and orientations for additional image capture during the subsequent drone flight.

\begin{figure*}
      \centering
      \includegraphics[width=1.0\textwidth]{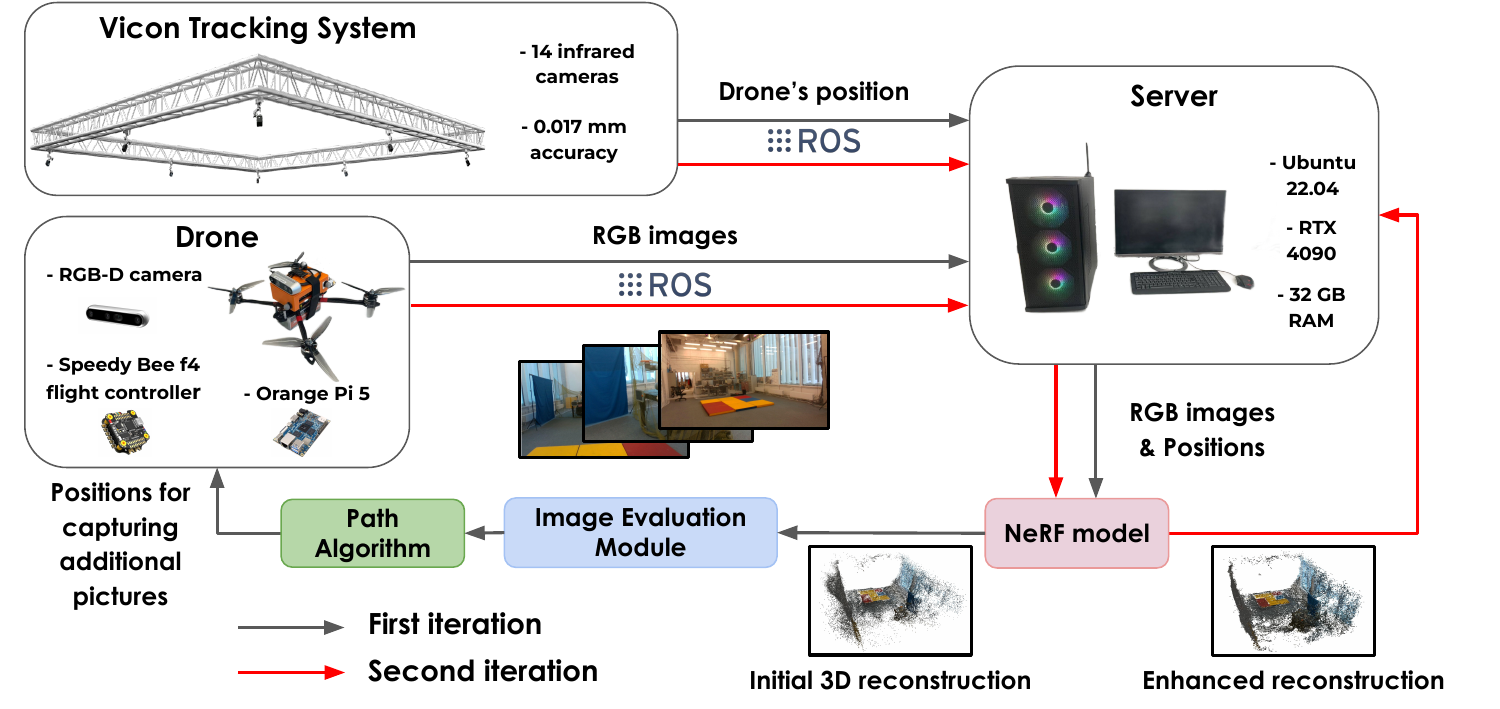}
      \caption{FlyNeRF system architecture. The grey pointers represent the initial dataset capture during the first iteration of the mission for 3D reconstruction. The red pointer denotes the second iteration dedicated to capturing additional images for enhancing reconstruction quality.}
      \label{arch}
   \end{figure*}

\section{Related Works}

\textbf{Neural Radiance Fields.} Implicit neural representations, particularly NeRF, pioneered by Mildenhall et al. \cite{mildenhall2020nerf}, offer a novel approach to generating complex 3D scenes. Leveraging neural networks to encode scene spatial layouts, NeRF represents a significant advancement, introducing a differentiable volume rendering loss. This innovation enables precise control over 3D scene reconstruction from 2D observations, achieving remarkable performance in novel view synthesis tasks. Despite the compelling properties and results demonstrated by vanilla NeRF \cite{mildenhall2020nerf}, the training process can be time-consuming. For instance, it typically necessitates approximately a day to train for a simple scene. This is due to the nature of the volume rendering, which requires a significant number of sample points to render an image. Subsequent research efforts have focused on improving the efficiency of the training and inference processes \cite{mueller2022instant, Reiser2021ICCV, Tancik_2023}, as well as overcoming the challenges of training from limited image data \cite{yu2020pixelnerf, kulhánek2022viewformer}. These advancements allow us to conduct 3D scene reconstruction on a single GPU.  

The 3D implicit representations are not limited to novel view synthesis. They have also been proposed for motion planning algorithms such as NFOMP \cite{9851532} and DNFOMP \cite{Katerishich_2023}, and for real-time SLAM algorithms such as IMAP \cite{Sucar_2021_ICCV}, NICE-SLAM, and iMODE \cite{zhu2022niceslam, 10161538}. These studies have shown the reliability of NeRF for our tasks related to navigation, localization, and path planning.  

% The first works \cite{Sucar_2021_ICCV} required RGB-D input and were used in limited settings. The following works such as \cite{zhu2022niceslam} introduced a hierarchical implicit representation of larger scenes, and the next iteration of this research \cite{zhu2023nicerslam} produced SLAM that requires only RGB input. The approaches of path planning combined with NeRF-based mapping include both ground and aerial vehicles, e.g. with a drone supporting the navigation of a quadruped robot in \cite{Zhura_2023}, however, the NFOMP for drones was not yet extensively explored.

\textbf{Autonomous scene exploration.} Scene exploration is a topic that has been studied in several works. It is usually aimed at maximizing coverage, with Frontier-Based Exploration (FBE) \cite{613851} being a common method. There are different variants of FBE \cite{6106778, topiwala2018frontier, 9560896}; however, the basic idea is to maintain a boundary between explored and unexplored areas and to sample points along this boundary. Recent studies have also investigated exploration using learning-based approaches \cite{DBLP:journals/corr/abs-2004-05155, modhe2023exploiting}. The study \cite{savinov2019episodic} introduces a novel approach to address the challenge of sparse rewards in reinforcement learning by leveraging agents' ability to generate rewards for encountering novel observations, similar to animals' curious behavior. In our work, we aim to improve the agent's ability of planning the optimal path during 3D reconstruction based on information about the location of points, where the image should be resampled.

\begin{figure*}
      \centering
      \includegraphics[width = 0.9\textwidth]{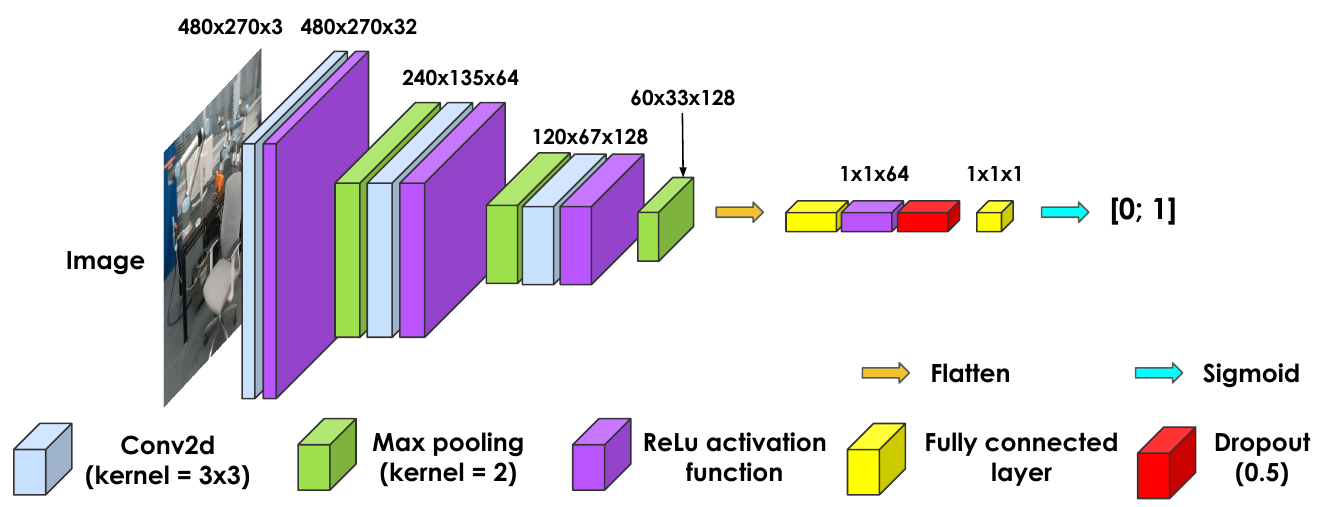}
      \caption{Image Evaluation Module architecture.}
      \label{cnn}
   \end{figure*}
   
\section{System Overview}

We developed a framework for aerial mapping enabling high-quality 3D reconstruction and environmental inspection. The FlyNeRF system architecture is presented in Fig.~\ref{arch}. The system integrates data capture by a UAV with a NeRF-based reconstruction pipeline. The environmental mapping mission includes two iterations: in the first iteration, the drone collects images of the environment, leading to the initial reconstruction. Following the assessment of the reconstruction's quality, additional positions for image capture are determined. The enhanced reconstruction is then generated using both the initial and additional data obtained from the two iterations.

In this work, we utilized a quadrotor with the Speedy Bee F4 flight controller and the single-board computer Orange Pi 5 for establishing the communication with the server. For image capture purposes the drone is equipped with an RGB-D RealSense D435i camera set to collect images in 1920x1080 resolution. The tracking of the drone's position during the flight was achieved through the Vicon Tracker system using 14 infrared ceiling-mounted cameras and reflective markers on the drone. This provided accurate spatial coordinates corresponding to each image capture location.

A fundamental software component used in our system is a NeRF-based model for 3D reconstruction of the environment. The quality assessment of the resulting reconstructions is made through renders represented as RGB images, from the NeRF model by the convolutional neural network (CNN) as an Image Evaluation Module. The NeRF model training and the utilization of the Image Evaluation Module are performed on a server equipped with the RTX 4090 GPU and 32 GB of RAM. 

The path planning algorithm is the key module for the adaptive improvement of the 3D reconstruction. It identifies regions with suboptimal rendering quality based on the Image Evaluation Module output, a probability, ranging from 0 for low-quality render to 1 for high-quality one. A probability threshold of lower than 0.7 categorizes areas as low quality. The algorithm then systematically interconnects areas that require additional image capturing, starting from the original drone position. 

In summary, the system serves as an automated tool for digitization of environments in 3D. Although our experiments focus on a particular setup, the modular design allows components to be adapted for a specific task. For instance, the UAV could be equipped with a different flight controller. The camera is interchangeable as well, to enable the selection of resolution and other attributes as required. Thus, the modular design allows tuning the system parameters to suit the goals of the reconstruction drone mission.

In the following subsections, we elaborate on certain system components that need a more detailed description. As the Image Evaluation Module requires training before integration into our system, we have dedicated a separate section to the dataset collection and its training process.
   
\subsection{3D Reconstruction}
To solve a 3D reconstruction task in an uncertain environment, NeRF technology is implemented in our system. NeRF has proved to be a highly effective approach for reconstructing 3D scenes, removing the need for expertise in 3D modeling and enabling rapid generation of high-fidelity results. Prior to NeRF model training, the preprocessing of the input data is required. After capturing images of the environment by UAV, the spatial position of each frame is recalculated from the positions obtained from the onboard drone log or the external motion caption system such as Vicon. 
 %The angle values are converted from quaternions to Euler angles, while the transition vector is passed for calculation without conversion. 
 Subsequently, we pass the obtained data captured from the drone to the NeRF model. With position information obtained, the images and corresponding coordinates in JSON format are provided as inputs for training the Nerfacto model \cite{Tancik_2023}, selected for its advantageous balance of reconstruction quality and efficiency. After model training is completed, the resulting point cloud and mesh outputs comprise the reconstructed 3D environment representation and could be used for tasks requiring a digital copy of the environment.

%The proposed NeRF integration enables high-precision 3D environmental mapping without requiring extensive modeling expertise, overcoming limitations of current reconstruction methodologies.

\subsection{Image Evaluation Module}\label{subsec:cnn_arch}

The Image Evaluation Module, a neural network, plays a crucial role in assessing the quality of renders generated from the NeRF-based 3D reconstruction. These renders, or novel views of the environment, are used as input for the Evaluation Module. Fig.~\ref{cnn} depicts the neural network's architecture, comprising three convolution layers for conducting a comprehensive analysis of the image's features and smoothly extracting them. The output is a probability representing the render's quality. A value close to 0 indicates poor quality, while a value closer to 1 signifies high-quality renders.

The overall neural network architecture is designed to progressively extract hierarchical features from the input data, facilitating effective representation learning for binary classification. Alongside convolutional layers for low-level feature capture with 3x3 kernels, we utilized the rectified linear unit (ReLU) activation to introduce non-linearity. Subsequent convolutional layers increase the network's capacity to distinguish complex spatial patterns, progressing from 32 to 64 and, finally, 128 output channels. The incorporation of max-pooling operations after each convolutional layer downsamples the spatial dimensions, preserving essential features while reducing computational complexity.

The next stage involves a flattening operation for a seamless transition from the tensor to a linear representation, preparing the data for fully connected layers. Further, we use two fully connected layers, each followed by ReLU activation and a dropout layer with a 0.5 dropout rate, to add regularisation and prevent overfitting during training. The final linear layer with a sigmoid activation is required for binary classification, producing a single output for decision-making.

%This architectural configuration is motivated by the need to capture hierarchical features in the input images, allowing the network to differentiate patterns and make binary classification decisions in subsequent stages of the learning process.

The binary classification problem was solved through the Binary Cross Entropy Loss. It is well-suited for problems where the objective is to classify input data into one of two classes, as is the case in binary classification scenarios. It specifically addresses scenarios like ours where each observation in the dataset is independent and does not require consideration of relationships between different instances.

\subsection{Classification Neural Network Training}

\begin{figure}
      \centering
      \includegraphics[scale=0.57]{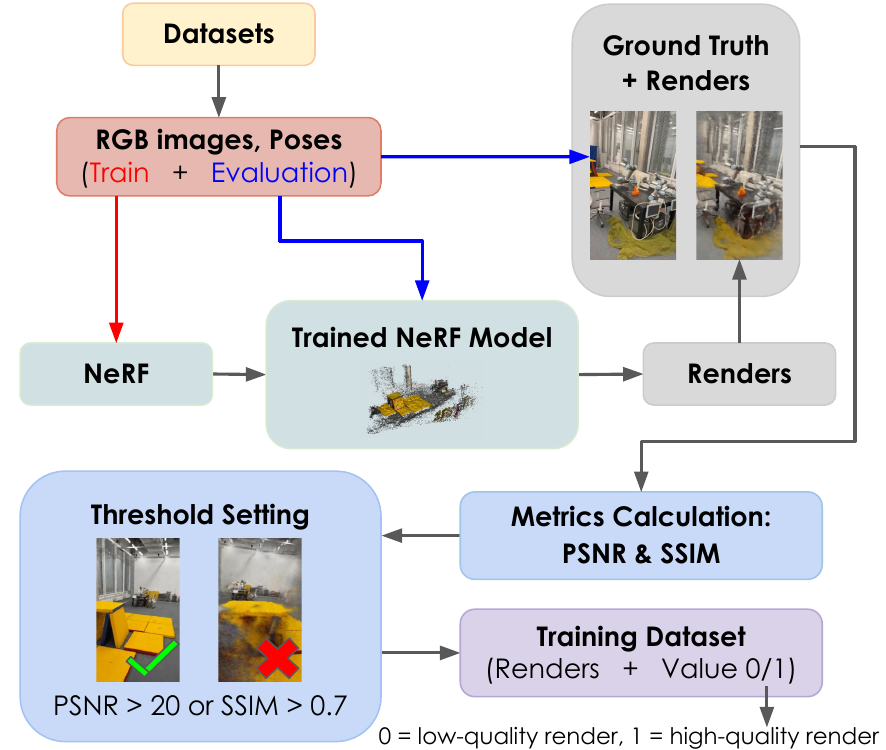}
      \caption{Dataset collection pipeline for the training of CNN-based image evaluation.}
      \label{dataset}
   \end{figure}

The training process of the proposed network included a dataset comprising renders from NeRF, each associated with a corresponding quality metric. Fig.~\ref{dataset} outlines the dataset collection and preparation process. To enhance the diversity of the initial RGB images for NeRF training, they were sourced from both various online repositories and through manual collection, ensuring a comprehensive representation.

Renders were acquired from predefined positions within the NeRF model. Ground truth images for these positions were available for reference-based image quality assessment during dataset preparation. Quality metrics were determined based on predefined thresholds for SSIM and PSNR values.

The choice of image quality assessment metrics is driven by specific objectives. PSNR assesses image fidelity by measuring levels of noise or distortion, while SSIM evaluates structural similarities, considering luminance, contrast, and structure. Thus, PSNR focuses on objective accuracy, and SSIM, in contrast, considers human visual perception and subjective visual appeal. The proposed pipeline utilizes both metrics for a comprehensive evaluation of image quality, addressing both objective and subjective aspects.

In conducting reference-based image quality assessment, an SSIM threshold of less than 0.7 and a PSNR value below 20 denoted renders as low-quality (0), while those exceeding these thresholds were considered high-quality (1). The determination of thresholds for metrics like SSIM or PSNR is application-dependent, and in our specific case, the values were chosen empirically, considering the image perception. The training dataset comprised an equal distribution of high and low-quality renders. The training process utilized a neural network configuration featuring 16.3 million parameters, evolving over 20 epochs to optimize the model's performance. The results of the proposed network are available in the Section~\ref{subsec:cnn_eval}.

\section{Experimental Evaluation}
To evaluate the proposed system, we conducted a series of experiments. The first step involved the drone flying along a predefined rectangular path, capturing RGB images and poses using the Vicon tracking system while facing the scene's center. Following the initial flight, collected data are transmitted to a server via the Robot Operating System (ROS) topic for NeRF model training and the generation of the initial environmental reconstruction. Subsequently, this model is used to render novel views from poses iteratively sampled within the predefined spatial bounds of the environment, with a certain increment between coordinates.

All NeRF-generated renders further undergo evaluation by the proposed in the \ref{subsec:cnn_arch} subsection Image Evaluation Module. As a result, we gather a list of probabilities indicating the likelihood of each render achieving high quality. Renders with probabilities below the predefined threshold are identified as suboptimal, requiring additional image capture from the corresponding poses. For this particular experiment, the threshold value was set to 0.7. The poses associated with low probabilities are compiled into a list intended for additional image capture.

To address abrupt probability changes in close spatial proximity, which could indicate the potential renderings within objects, such abrupt alterations are disregarded. Only smoothly declining quality regions are identified as requiring further image capture. The resulting list of additional image capture positions is afterward transmitted to the drone, initiating the second iteration of the mission.

In the second phase, the drone navigates to the specified poses and captures additional images. The resulting data are added to the initially collected dataset. This expanded dataset is then utilized for training the enhanced NeRF model, leading to an improved 3D reconstruction of the environment. 

\begin{comment}
\begin{algorithm}
\caption{FlyNeRF system algorithm}\label{alg:experim}
\begin{algorithmic}[1]
\Procedure{DataCollection}{}
    \State Drone follows a predefined path, capturing images and spatial coordinates.
    \State Dataset includes images and ground truth coordinates.
\EndProcedure

\Procedure{NeRFReconstruction}{}
    \State Initial 3D reconstruction using Neural Radiance Fields (NeRF).
\EndProcedure

\Procedure{RenderQualityAssessment}{}
    \State Classification neural network assesses render quality.
    \State Training on a curated dataset with labeled render quality.
\EndProcedure

\Procedure{AdaptiveRefinement}{}
    \State Algorithm identifies areas of suboptimal render quality.
    \State Drone revisits areas, captures additional images for refinement.
    \State Secondary NeRF model enhances overall reconstruction quality.
\EndProcedure
\end{algorithmic}
\end{algorithm}

\begin{algorithm}
\caption{Euclid’s algorithm}\label{alg:euclid}
\begin{algorithmic}[1]
\Procedure{Euclid}{$a,b$}\Comment{The g.c.d. of a and b}
\State $r\gets a\bmod b$
\While{$r\not=0$}\Comment{We have the answer if r is 0}
\State $a\gets b$
\State $b\gets r$
\State $r\gets a\bmod b$
\EndWhile\label{euclidendwhile}
\State \textbf{return} $b$\Comment{The gcd is b}
\EndProcedure
\end{algorithmic}
\end{algorithm}
\end{comment}

The following sections are dedicated to demonstrating the evaluation of the reconstruction quality and performance of the neural network proposed for image quality assessment.

\subsection{Image Evaluation Module}\label{subsec:cnn_eval}

\begin{figure}
    \vspace{5pt}
    \captionsetup[subfloat]{skip=0.5pt}
    \centering
    \hfill
    \begin{subfigure}{0.23\textwidth}
        \centering
        \includegraphics[width=\linewidth]{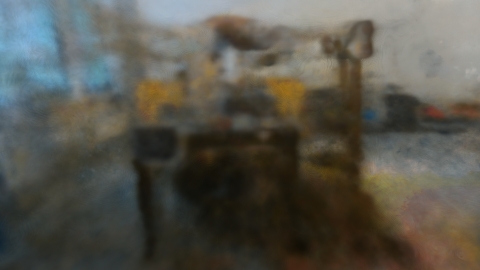}
        \captionsetup{labelformat=empty}
        \caption{(a)}
        \label{subfig:a}
    \end{subfigure}
    \hfill
        \begin{subfigure}{0.23\textwidth}
        \centering
        \includegraphics[width=\linewidth]{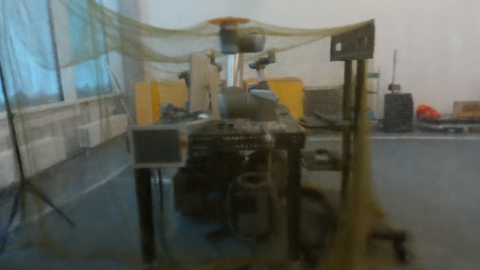}
        \captionsetup{labelformat=empty}
        \caption{(d)}
        \label{subfig:d}
    \end{subfigure}
    \\
    \vspace{4pt}
    \hfill
    \begin{subfigure}{0.23\textwidth}
        \centering
        \includegraphics[width=\linewidth]{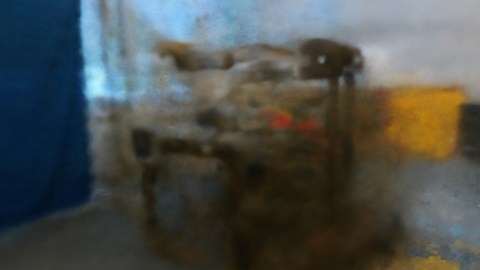}
        \captionsetup{labelformat=empty}
        \caption{(b)}
        \label{subfig:b}
    \end{subfigure}
    \hfill
        \begin{subfigure}{0.23\textwidth}
        \centering
        \includegraphics[width=\linewidth]{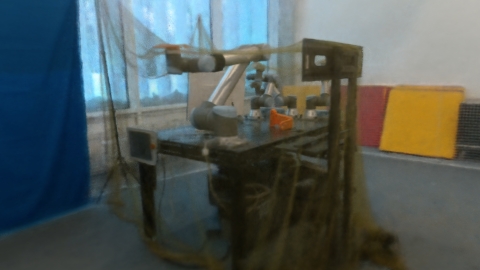}
        \captionsetup{labelformat=empty}
        \caption{(e)}
        \label{subfig:e}
    \end{subfigure}
    \\
    \vspace{4pt}
    \hfill
    \begin{subfigure}{0.23\textwidth}
        \centering
        \includegraphics[width=\linewidth]{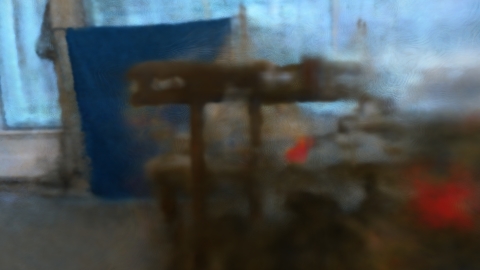}
        \captionsetup{labelformat=empty}
        \caption{(c)}
        \label{subfig:c}
    \end{subfigure}
    \hfill
    \begin{subfigure}{0.23\textwidth}
        \centering
        \includegraphics[width=\linewidth]{images/e.jpg}
        \captionsetup{labelformat=empty}
        \caption{(f)}
        \label{subfig:f}
    \end{subfigure}

    \caption{Render comparison after (a-c) the first and (d-f) the second iteration of the mission.}
    \label{fig:compar}
\end{figure}

 Visual quality comparison among certain renders obtained during the mission is provided in Fig.~\ref{fig:compar}. Images (a)-(c) depict renders received from the NeRF model after the initial reconstruction, whereas images (d)-(f) illustrate renders from the same positions following the second iteration of the mission and additional image capture. Additionally, the comparison for image evaluation module evaluation is presented in Table~\ref{tab:comp}.  

Table \ref{tab:comp} illustrates the image evaluation module's capability to evaluate render quality in accordance with the metrics. Renders from the NeRF model after the initial flight are identified as having mediocre quality, with SSIM and PSNR values below the predefined thresholds of 0.7 and 20, respectively. In contrast, the second flight results suggest renders are likely of high quality with a probability exceeding 0.84. Moreover, metrics show significant improvement, emphasizing the evaluation module's capacity to detect enhancements in render quality. 

The proposed neural network achieves an accuracy of 0.97 and a ROC AUC score of 0.99, indicating its excellent ability to differentiate between image classes.

\begin{table}
    \centering
    \caption{Render comparison}
   \begin{tabular}{|c|c|c|c|}
        \hline
        \multicolumn{4}{|c|}{\textbf{First Iteration}} \\
        \hline 
        & \multirow{2}{*}{\textbf{Predicted Quality}} & \multicolumn{2}{|c|}{\textbf{Metrics}} \\
        \cline{3-4}
        & & \textbf{PSNR} & \textbf{SSIM} \\
        \hline
        a) & 0.60 & 16.01 & 0.63 \\
        \hline
        b) & 0.39 & 15.33 & 0.68 \\
        \hline
        c) & 0.003 & 15.37 & 0.56 \\
        \hline
        \hline
        \multicolumn{4}{|c|}{\textbf{Second Iteration}} \\
        \hline 
        & \multirow{2}{*}{\textbf{Predicted Quality}} & \multicolumn{2}{|c|}{\textbf{Metrics}} \\
        \cline{3-4}
        & & \textbf{PSNR} & \textbf{SSIM} \\
        \hline
        d) & 0.99 & 20.23 & 0.84 \\
        \hline
        e) & 0.99 & 18.24 & 0.81 \\
        \hline
        f) & 0.84 & 17.01 & 0.73 \\
        \hline
    \end{tabular}
    
    \label{tab:comp}
\end{table}

\subsection{Reconstruction Quality}

\begin{figure}
    \centering
    \begin{subfigure}{0.23\textwidth}
        \centering
        \includegraphics[width=\linewidth]{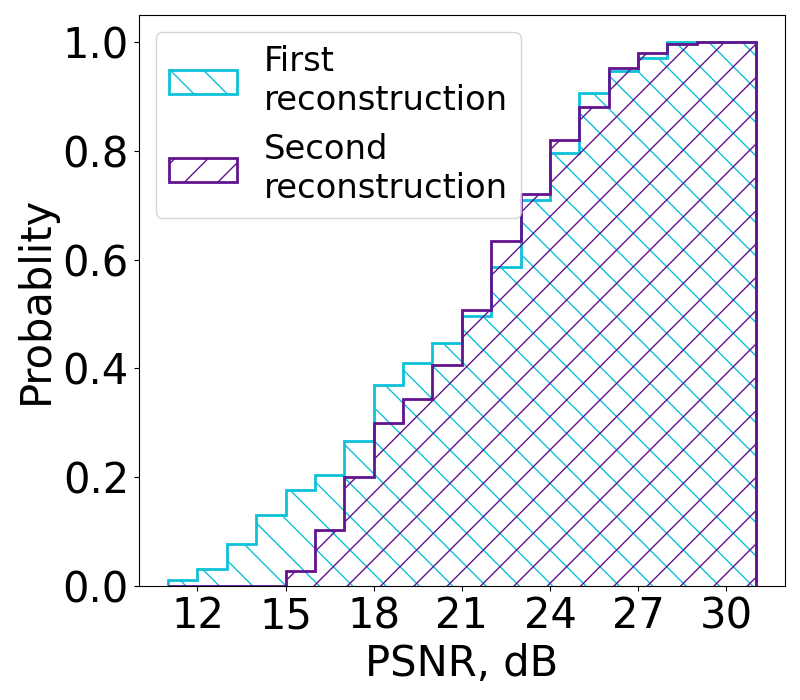}
        \caption{}
        \label{subfig:SSIM}
    \end{subfigure}
    \begin{subfigure}{0.23\textwidth}
        \centering
        \includegraphics[width=\linewidth]{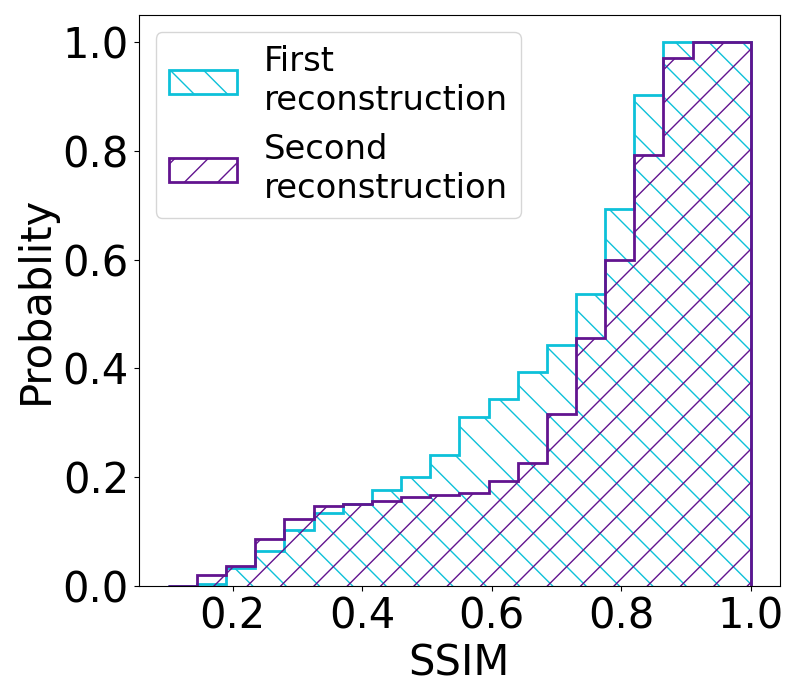}
        \caption{}
        \label{subfig:PSNR}
    \end{subfigure}
    \caption{Cumulative distribution functions for (a) SSIM and (b) PSNR for the first and second 3D reconstruction.}
    \label{fig:metrics}
\end{figure}

To evaluate reconstruction quality, we computed PSNR and SSIM values for renders from both iterations, presenting them as cumulative distribution functions (Fig.~\ref{subfig:SSIM}).
The SSIM values for the second reconstruction show a quicker rise, reaching higher similarity levels, up to 0.93. In comparison, the first reconstruction has a larger share of images falling within the 0.4--0.7 range, and the maximum similarity achieved is slightly lower. However, in the second iteration, a minor fraction of renders have lower SSIM values compared to those from the first flight. The aforementioned observation could be due to an increased percentage of blurry images captured during the flight and subsequently utilized in NeRF model training. This suggests the potential need for an input filter for images used in NeRF training.

The PSNR values remain mostly consistent on average, shown in Fig.~\ref{subfig:PSNR}. However, there's a noticeable difference in the PSNR values lower than 15 dB. This lower range area is the focus of our system, which is designed to improve the quality of poorly reconstructed areas in the environment. 

A closer analysis (Table~\ref{tab:quantile}) focuses on the lowest quartile (Q1) and other lower probabilities quantiles, specifically 0.1 and 0.05. These quantiles indicate the value below which 25{\%}, 10{\%}, and 5{\%} of the PSNR values fall, accordingly. 
The PSNR values increased across all quantiles between the first and second 3D reconstructions, indicating improved fidelity and reduced noise in the final reconstruction. Notably, we observe a more evident increase in PSNR as the quantile decreases. This indicates the successful achievement of our goal to enhance the quality of poorly reconstructed areas.   
% \begin{figure}[h!]
%     \centering
%     \includegraphics[scale=0.35]{images/3.png}
%     \caption{Quantiles for the PSNR values.}
%     \label{fig:quantiles}
% \end{figure}

\begin{table}[]
\centering
\caption{PSNR Quantiles}
\begin{tabular}{cccll}
\cline{1-3}
\multicolumn{1}{|c|}{\multirow{2}{*}{\textbf{Quantile}}} & \multicolumn{2}{c|}{\textbf{Iteration}}                        &  &  \\ \cline{2-3}
\multicolumn{1}{|c|}{}                          & \multicolumn{1}{c|}{\textbf{I}}    & \multicolumn{1}{c|}{\textbf{II}}   &  &  \\ \cline{1-3}
\multicolumn{1}{|c|}{0.25} & \multicolumn{1}{c|}{17.89} & \multicolumn{1}{c|}{18.41} &  &  \\ \cline{1-3}
\multicolumn{1}{|c|}{0.1}                       & \multicolumn{1}{c|}{14.51} & \multicolumn{1}{c|}{16.98}   &  &  \\ \cline{1-3}
\multicolumn{1}{|c|}{0.05}                      & \multicolumn{1}{c|}{13.81} & \multicolumn{1}{c|}{16.42} &  &  \\ \cline{1-3}
\multicolumn{1}{l}{}                            & \multicolumn{1}{l}{}      & \multicolumn{1}{l}{}      &  & 
\end{tabular}
\label{tab:quantile}
\end{table}

\begin{figure}[h!]
    \centering
    \begin{subfigure}{0.24\textwidth}
        \centering
        \includegraphics[width=\linewidth]{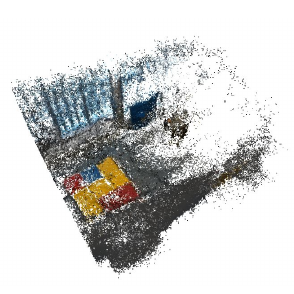}
        \caption{}
        \label{subfig:rec1}
    \end{subfigure}%
    \begin{subfigure}{0.24\textwidth}
        \centering
        \includegraphics[width=\linewidth]{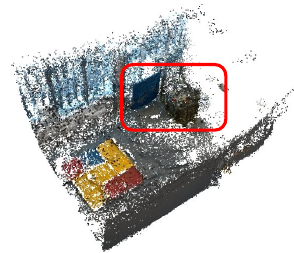}
        \caption{}
        \label{subfig:rec2}
    \end{subfigure}
    \caption{3D reconstruction represented as a point cloud after (a) the first and (b) the second flight.}
    \label{fig:rec}
\end{figure}

Fig.~\ref{fig:rec} illustrates two 3D reconstructions depicted as point clouds: one obtained from the NeRF model after the initial flight and another from the second iteration of the mission. The red zone in Fig.~\ref{subfig:rec2} highlights the region where additional image capture occurred. The object's outline becomes notably more distinct after the supplementary image capture, making it more suitable for subsequent applications.

\section{Conclusion and Future Work}
In this work, we introduced FlyNeRF, a system designed for efficient mapping and improvement of 3D reconstructions in unknown environments using UAVs, and a NeRF-based model for reconstruction. Our methodology employs a convolutional neural network to assess the quality of renders derived from 3D reconstructions, providing a probability indicating the image's quality. The proposed network achieves an accuracy of 0.97 and a ROC AUC score of 0.99. The results from the network enable efficient position determination for additional image capture by the drone. The second reconstruction demonstrates significant quality enhancement, supported by PSNR and SSIM metrics. In a series of uniform experiments involving two sequential flights, we observe an average improvement of 2.5 dB in PSNR for the 10{\%} quantile, indicating successful enhancement of low-quality areas in the reconstruction.

The FlyNeRF system offers potential applications across diverse domains. Beyond its core role in 3D mapping and reconstruction, its ability to enhance poorly reconstructed areas extends its utility to environmental monitoring, urban planning, and disaster response. Additionally, our system could provide efficient data collection for infrastructure inspection, agricultural monitoring, and other scenarios requiring detailed and accurate 3D reconstructions.

Future work involves implementing a quality filter for collected images used for NeRF model training in order to achieve improved results with fewer images. Our other objective is to eliminate the dependency on the Vicon for capturing drone positions, thus enhancing the flexibility of our system and enabling its use in outdoor environments. Additionally, we plan to explore the path planning aspect, evaluating different strategies for collecting positions for additional image capture instead of the current subsequent one. Furthermore, validation of our approach in a simulation environment could offer faster and more flexible evaluations compared to real-life experiments.

\section*{Acknowledgements} 
Research reported in this publication was financially supported by the Russian Science Foundation grant No. 24-41-02039.

\bibliographystyle{IEEEbib}
\bibliography{sources}

\begin{thebibliography}{10}

\bibitem{mutti2022unsupervised}
Mirco Mutti, Mattia Mancassola, and Marcello Restelli,
\newblock ``Unsupervised reinforcement learning in multiple environments,''
\newblock in {\em Proceedings of the AAAI Conference on Artificial Intelligence}, 2022, vol.~36, pp. 7850--7858.

\bibitem{savinov2019episodic}
Nikolay Savinov, Anton Raichuk, Raphaël Marinier, Damien Vincent, Marc Pollefeys, Timothy Lillicrap, and Sylvain Gelly,
\newblock ``Episodic curiosity through reachability,'' 2019.

\bibitem{ramakrishnan2022poni}
Santhosh~K. Ramakrishnan, Devendra~Singh Chaplot, Ziad Al-Halah, Jitendra Malik, and Kristen Grauman,
\newblock ``Poni: Potential functions for objectgoal navigation with interaction-free learning,''
\newblock in {\em Computer Vision and Pattern Recognition (CVPR), 2022 IEEE Conference on}. IEEE, 2022.

\bibitem{liang2021sscnav}
Yiqing Liang, Boyuan Chen, and Shuran Song,
\newblock ``Sscnav: Confidence-aware semantic scene completion for visual semantic navigation,''
\newblock in {\em Proc. of The International Conference in Robotics and Automation (ICRA)}, 2021.

\bibitem{krantz2020navgraph}
Jacob Krantz, Erik Wijmans, Arjun Majumdar, Dhruv Batra, and Stefan Lee,
\newblock ``Beyond the nav-graph: Vision-and-language navigation in continuous environments,'' 2020.

\bibitem{min2021film}
So~Yeon Min, Devendra~Singh Chaplot, Pradeep Ravikumar, Yonatan Bisk, and Ruslan Salakhutdinov,
\newblock ``Film: Following instructions in language with modular methods,'' 2021.

\bibitem{nguyen2021look}
Van-Quang Nguyen, Masanori Suganuma, and Takayuki Okatani,
\newblock ``Look wide and interpret twice: Improving performance on interactive instruction-following tasks,'' 2021.

\bibitem{mildenhall2020nerf}
Ben Mildenhall, Pratul~P. Srinivasan, Matthew Tancik, Jonathan~T. Barron, Ravi Ramamoorthi, and Ren Ng,
\newblock ``Nerf: Representing scenes as neural radiance fields for view synthesis,''
\newblock in {\em ECCV}, 2020.

\bibitem{mueller2022instant}
Thomas M\"uller, Alex Evans, Christoph Schied, and Alexander Keller,
\newblock ``Instant neural graphics primitives with a multiresolution hash encoding,''
\newblock {\em ACM Trans. Graph.}, vol. 41, no. 4, pp. 102:1--102:15, July 2022.

\bibitem{Reiser2021ICCV}
Christian Reiser, Songyou Peng, Yiyi Liao, and Andreas Geiger,
\newblock ``Kilonerf: Speeding up neural radiance fields with thousands of tiny mlps,''
\newblock in {\em International Conference on Computer Vision (ICCV)}, 2021.

\bibitem{Tancik_2023}
Matthew Tancik, Ethan Weber, Evonne Ng, Ruilong Li, Brent Yi, Terrance Wang, Alexander Kristoffersen, Jake Austin, Kamyar Salahi, Abhik Ahuja, David Mcallister, Justin Kerr, and Angjoo Kanazawa,
\newblock ``Nerfstudio: A modular framework for neural radiance field development,''
\newblock in {\em Special Interest Group on Computer Graphics and Interactive Techniques Conference Conference Proceedings}. July 2023, SIGGRAPH ’23, ACM.

\bibitem{yu2020pixelnerf}
Alex Yu, Vickie Ye, Matthew Tancik, and Angjoo Kanazawa,
\newblock ``{pixelNeRF}: Neural radiance fields from one or few images,''
\newblock in {\em CVPR}, 2021.

\bibitem{kulhánek2022viewformer}
Jonáš Kulhánek, Erik Derner, Torsten Sattler, and Robert Babuška,
\newblock ``Viewformer: Nerf-free neural rendering from few images using transformers,'' 2022.

\bibitem{9851532}
Mikhail Kurenkov, Andrei Potapov, Alena Savinykh, Evgeny Yudin, Evgeny Kruzhkov, Pavel Karpyshev, and Dzmitry Tsetserukou,
\newblock ``Nfomp: Neural field for optimal motion planner of differential drive robots with nonholonomic constraints,''
\newblock {\em IEEE Robotics and Automation Letters}, vol. 7, no. 4, pp. 10991--10998, 2022.

\bibitem{Katerishich_2023}
Maksim Katerishich, Mikhail Kurenkov, Sausar Karaf, Artem Nenashev, and Dzmitry Tsetserukou,
\newblock ``Dnfomp: Dynamic neural field optimal motion planner for navigation of autonomous robots in cluttered environment,''
\newblock in {\em 2023 IEEE International Conference on Systems, Man, and Cybernetics (SMC)}, 2023, pp. 1984--1989.

\bibitem{Sucar_2021_ICCV}
Edgar Sucar, Shikun Liu, Joseph Ortiz, and Andrew~J. Davison,
\newblock ``imap: Implicit mapping and positioning in real-time,''
\newblock in {\em Proceedings of the IEEE/CVF International Conference on Computer Vision (ICCV)}, October 2021, pp. 6229--6238.

\bibitem{zhu2022niceslam}
Zihan Zhu, Songyou Peng, Viktor Larsson, Weiwei Xu, Hujun Bao, Zhaopeng Cui, Martin~R. Oswald, and Marc Pollefeys,
\newblock ``Nice-slam: Neural implicit scalable encoding for slam,'' 2022.

\bibitem{10161538}
Hidenobu Matsuki, Edgar Sucar, Tristan Laidow, Kentaro Wada, Raluca Scona, and Andrew~J. Davison,
\newblock ``imode:real-time incremental monocular dense mapping using neural field,''
\newblock in {\em 2023 IEEE International Conference on Robotics and Automation (ICRA)}, 2023, pp. 4171--4177.

\bibitem{613851}
B.~Yamauchi,
\newblock ``A frontier-based approach for autonomous exploration,''
\newblock in {\em Proceedings 1997 IEEE International Symposium on Computational Intelligence in Robotics and Automation CIRA'97. 'Towards New Computational Principles for Robotics and Automation'}, 1997, pp. 146--151.

\bibitem{6106778}
Christian Dornhege and Alexander Kleiner,
\newblock ``A frontier-void-based approach for autonomous exploration in 3d,''
\newblock in {\em 2011 IEEE International Symposium on Safety, Security, and Rescue Robotics}, 2011, pp. 351--356.

\bibitem{topiwala2018frontier}
Anirudh Topiwala, Pranav Inani, and Abhishek Kathpal,
\newblock ``Frontier based exploration for autonomous robot,'' 2018.

\bibitem{9560896}
Anthony Brunel, Amine Bourki, Cédric Demonceaux, and Olivier Strauss,
\newblock ``Splatplanner: Efficient autonomous exploration via permutohedral frontier filtering,''
\newblock in {\em 2021 IEEE International Conference on Robotics and Automation (ICRA)}, 2021, pp. 608--615.

\bibitem{DBLP:journals/corr/abs-2004-05155}
Devendra~Singh Chaplot, Dhiraj Gandhi, Saurabh Gupta, Abhinav Gupta, and Ruslan Salakhutdinov,
\newblock ``Learning to explore using active neural {SLAM},''
\newblock {\em CoRR}, vol. abs/2004.05155, 2020.

\bibitem{modhe2023exploiting}
Nirbhay Modhe, Qiaozi Gao, Ashwin Kalyan, Dhruv Batra, Govind Thattai, and Gaurav Sukhatme,
\newblock ``Exploiting generalization in offline reinforcement learning via unseen state augmentations,'' 2023.

\end{thebibliography}

\end{document}